\documentclass[%
 reprint,
 superscriptaddress,
nobibnotes,
 amsmath,amssymb,
 aps,
floatfix,
]{revtex4-2}

\usepackage{graphicx}% Include figure files
\usepackage{dcolumn}% Align table columns on decimal point
\usepackage{bm}% bold math
\usepackage{enumitem}

\usepackage{physics}
\usepackage{hyperref}% add hypertext capabilities
\usepackage{xcolor}

\usepackage{comment}

\begin{document}

%\preprint{APS/123-QED}

\title{ToPolyAgent: AI Agents for Coarse-Grained Topological Polymer Simulations}

\author{Lijie Ding}
\email{dingl1@ornl.gov}
\affiliation{Neutron Scattering Division, Oak Ridge National Laboratory, Oak Ridge, TN 37831, USA}
\author{Jan-Michael Carrillo}
\affiliation{Center for Nanophase Materials Sciences, Oak Ridge National Laboratory, Oak Ridge, TN 37831, USA}
\author{Changwoo Do}
\email{doc1@ornl.gov}
\affiliation{Neutron Scattering Division, Oak Ridge National Laboratory, Oak Ridge, TN 37831, USA}

\date{\today}% It is always \today, today,
             %  but any date may be explicitly specified

\begin{abstract}
We introduce ToPolyAgent, a multi-agent AI framework for performing coarse-grained molecular dynamics (MD) simulations of topological polymers through natural language instructions. By integrating large language models (LLMs) with domain-specific computational tools, ToPolyAgent supports both interactive and autonomous simulation workflows across diverse polymer architectures, including linear, ring, brush, and star polymers, as well as dendrimers. The system consists of four LLM-powered agents: a Config Agent for generating initial polymer-solvent configurations, a Simulation Agent for executing LAMMPS-based MD simulations and conformational analyses, a Report Agent for compiling markdown reports, and a Workflow Agent for streamlined autonomous operations. Interactive mode incorporates user feedback loops for iterative refinements, while autonomous mode enables end-to-end task execution from detailed prompts. We demonstrate ToPolyAgent's versatility through case studies involving diverse polymer architectures under varying solvent condition, thermostats, and simulation lengths. Furthermore, we highlight its potential as a research assistant by directing it to investigate the effect of interaction parameters on the linear polymer conformation, and the influence of grafting density on the persistence length of the brush polymer. By coupling natural language interfaces with rigorous simulation tools, ToPolyAgent lowers barriers to complex computational workflows and advances AI-driven materials discovery in polymer science. It lays the foundation for autonomous and extensible multi-agent scientific research ecosystems.
\end{abstract}
%\keywords{Suggested keywords}%Use showkeys class option if keyword
                              %display desired
\maketitle

%\tableofcontents

\section{Introduction}

% 1. topological polymers
Topological polymers\cite{tezuka2002topological,yamamoto2011topological,binder1995monte}, characterized by their diverse architectures such as linear\cite{casassa1969equilibrium,baschnagel1991construction}, ring\cite{braams2006short,habershon2013ring}, brush\cite{murat1989structure,feng2018polymer,milner1988theory,milner1991polymer}, star\cite{ren2016star,grest1987structure}, and dendrimer\cite{tomalia1995dendrimer} structures, exhibit unique physical properties that make them critical in applications ranging from drug delivery\cite{kesharwani2014dendrimer,sung2020recent,liechty2010polymers,qiu2006polymer,pillai2001polymers} to advanced materials\cite{brazel2012fundamental,tan2020recent}. These properties arise from their complex molecular configurations, which influence their conformational dynamics, phase behavior, and interactions with solvent\cite{fixman1970polymer,koningsveld2001polymer,watzlawek1999phase,bates1991polymer,muller2002phase,liu2022machine}. Understanding these behaviors often requires molecular dynamics (MD) simulations\cite{allen2017computer,binder1995monte,hollingsworth2018molecular,hansson2002molecular}, which provide detailed insights into the polymer structure and dynamics at scales from atomistic to coarse-grained\cite{ding2024off,muller2002coarse,zhao2017overview,dunweg1993molecular}. However, despite the development of professional simulation software such as LAMMPS\cite{thompson2022lammps}, GROMACS\cite{van2005gromacs}, AMBER\cite{salomon2013overview}, and Desmond\cite{chow2008desmond}, the computational complexity of simulating topological polymers, even in coarse-grained models, still demands significant expertise, limiting their accessibility to researchers without extensive computational backgrounds.

% 2. LLMs and agentic AI
Recent advances in large language models (LLMs)\cite{brown2020language,radford2019language,naveed2025comprehensive} and agentic AI\cite{acharya2025agentic,gridach2025agentic} have opened new avenues for automating and simplifying complex scientific workflows\cite{boiko2023autonomous}. LLMs, with their ability to process and generate natural language, can interpret user instructions, orchestrate tasks, and integrate domain-specific tools to perform specialized computations\cite{shen2024llm,masterman2024landscape}. Agentic AI systems, composed of multiple AI agents with distinct roles, further enhance this capability by enabling collaborative and autonomous task execution. These developments have the potential to transform computational materials science by making advanced simulations more intuitive and efficient, while greatly lowering the barrier to using professional computational tools like MD software. The application of AI agent started with software development\cite{yang2024swe, liu2024marscode}, and has been quickly expanding to various areas of scientific research\cite{schmidgall2025agent,mendible2025dynamate,ding2025sasagent}. Despite these advancement, we still lack an agentic AI system for carrying out in-depth simulation work with sufficient generality to cover a wide range of topological polymer simulation.

% 3. what does this work do
In this work, we address the aforementioned challenges by introducing ToPolyAgent, a multi-agent AI framework designed to perform coarse-grained molecular dynamics (MD) simulations of topological polymers through natural language interfaces. The coarse-grained MD simulation workflow is based on a tutorial supported by the U.S. National Science Foundation (NSF)~\cite{rpi_md_tutorial,giedt2018eager}. ToPolyAgent integrates large language models (LLMs) with domain-specific computational tools to support simulations of diverse polymer architectures under varying solvent conditions and simulation parameters. The system operates in two modes: an interactive mode, which incorporates user feedback for iterative refinement, and an autonomous mode, which performs end-to-end simulations from detailed prompts.

We demonstrate ToPolyAgent’s capabilities through case studies that highlight its potential as a research assistant, including investigations of the effect of solvent quality on linear polymer conformation and the influence of grafting density on the persistence length of brush polymer backbones. By bridging natural language processing with rigorous computational methods, ToPolyAgent lowers the barrier to performing complex polymer simulations and establishes a foundation for scalable, AI-driven materials research.

% structure of this work
The rest of this paper is organized as follows: in Sec~\ref{sec:method}, we describe the system design of ToPolyAgent, and provide details on the tools been used and technical detail on the MD simulation. Sec.~\ref{sec:result} presents a series of examples and use case of the ToPolyAgent simulating different types of topological polymer under various conditions, while demonstrating the capability of ToPolyAgent as a research assistant. Finally, we summarize this work and discuss potential future direction in Sec.~\ref{sec:summary}.

\begin{figure}[!t]
    \centering
    \includegraphics[width=1\linewidth]{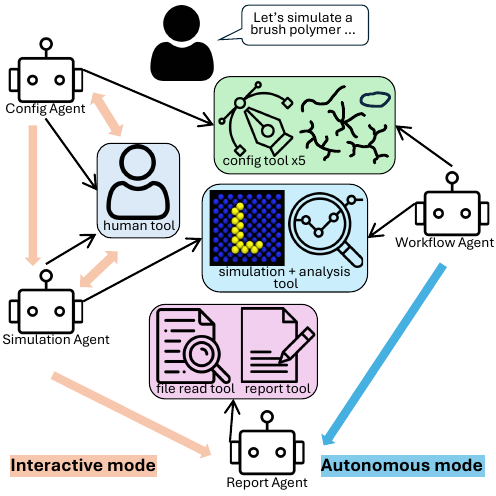}
    \caption{Overview of the ToPolyAgent workflow. The system operates in two modes: interactive and autonomous. Four sets of tools are assigned to four distinct agents. These agents execute molecular dynamics simulations of various topological polymers under different conditions based on user instructions.}
    \label{fig:workflow}
\end{figure}

\section{Method}
\label{sec:method}

\subsection{System Design}

ToPolyAgent is designed not only to perform coarse-grained MD simulations for various types of topological polymers directly from detailed prompts but also able to interact with users and adjust simulation settings based on their feedback. We implement this as a multi-agent system with two operational modes: interactive mode and autonomous mode. Fig.~\ref{fig:workflow} illustrates the overall system design. Four distinct agents support these two modes in conducting MD simulations for topological polymers. The interactive mode is supported by the Config Agent, Simulation Agent, and Report Agent, while the autonomous mode relies on the Workflow Agent and Report Agent.

In the interactive mode, the first agent to interact with the user is the Config Agent, whose task is to generate the initial configuration data file containing the positions and connectivities of all polymer beads, as well as the initial positions of all solvent particles. The Config Agent employs six different tools to perform its tasks. Five of these tools are dedicated to generating five types of topological polymers — linear, ring, brush, star, and dendrimer — each with specific input parameters that control properties such as size, shape, and solvent concentration. The remaining tool, known as the human tool, prompts the user for additional feedback so that the Config Agent can iteratively modify the system configuration until the user is satisfied and agrees to proceed to the next phase.

The second phase of the interactive mode is handled by the Simulation Agent, which interprets user prompts and runs molecular dynamics (MD) simulations for the previously generated polymer configurations using appropriate parameters. In addition, the Simulation Agent performs data analysis on the simulation output and presents figures for user evaluation. Similar to the Config Agent, the Simulation Agent also has access to the human tool and will request user feedback to adjust and rerun simulations as needed.

Once the user approves the simulation results, the interactive mode proceeds to the third phase, in which all results --- including simulation data, parameters, and interaction logs --- are compiled by the Report Agent. The Report Agent uses tools to read logs and files, then generates a comprehensive simulation report in markdown format.

In the autonomous mode, the agents expect the user to provide more detailed prompts so that the simulations can be carried out with greater specificity. The main difference between the autonomous and interactive modes is the absence of a human feedback loop in the former. In this mode, the Workflow Agent is responsible for generating the initial system configuration, performing the MD simulation of the polymer–solvent system, and conducting data analysis. The resulting outputs are then passed to the same Report Agent, which compiles them into a final report.

In practice, we use CrewAI to orchastrate the multi-agent system and facilitate agent-to-agent collaboration, as well as to maintain the agents' memory systems. The LLM is powered by OpenRouter's API, which enables flexible selection of the underlying model. And without loss of generality, we use OpenAI's GPT-4o-mini\cite{achiam2023gpt} for the results presented in this study.

\subsection{Tools for Agents}

To enable the different agents to accomplish their designated tasks, we develope toolsets tailored to each agent's tasks. The configuration generation toolset creates the initial configurations for LAMMPS to read, simulate, and visualize through plots. The simulation toolset runs the MD simulations and performs conformational analysis using the dump files produced by LAMMPS. The human tool connects the agents to the user via command-line inputs. Finally, the report toolset allows agents to read the files generated during the simulations and compile reports in markdown format.

%need some reference for all kinds of polymer here (maybe, if not in intro)
For system configuration generation, five tools were developed, each corresponding to a specific type of topological polymer. These tools read predefined input parameters from docstrings, generate the polymer configurations based on user instructions, add solvent to the simulation box, save the resulting LAMMPS data file, and visualize the configurations. All tools require the simulation box size $B$ and solvent density $n_s$ as inputs, with additional topology-specific parameters. \textit{GenerateLinearPolymer} and \textit{GenerateRingPolymer} require the chain length $N$. \textit{GenerateBrushPolymer} requires the backbone length $N_b$, grafting density $\sigma_g$, and side chain length $N_s$. \textit{GenerateStarPolymer} requires the arm length $N_a$ and number of arms $m$. \textit{GenerateDendrimer} requires the number of generations $G$, branching factor $b$, and spacer length $N_s$.

For the simulation tools set, the \textit{runLAMMPS} tool executes the MD simulation using the LAMMPS data file generated by the Config Agent.  It  takes as input the interaction parameters: polymer-polymer $\epsilon_{pp}$, solvent-solvent $\epsilon_{ss}$, and polymer-solvent $\epsilon_{sp}$, along with the thermostat type (Langevin or Nose-Hoover) and the number of simulation steps. Details of the MD simulations are provided in the following subsection. Additionally, the \textit{ComformationAnalysis} tool analyzes the dump files produced by the LAMMPS simulations. 

The following characteristic variables of polymer conformation are calculated~\cite{de1979scaling,rubinstein2003polymer}: the square of the radius of gyration, $R_g^2 = \left< (\vb{r}_i - \vb{r}_c)^2 \right>_i$, where $\left< \dots \right>_i$ denotes the average over all beads on the polymer, and $\vb{r}_c = \left< \vb{r}_i \right>_i$ is the center of mass of the polymer. The mean square displacement (MSD) curve is given by $\text{MSD}(\Delta t) = \left< |\vb{r}_c(t + \Delta t) - \vb{r}_c(t)|^2 \right>_t$, where $\left< \dots \right>_t$ denotes the time average, and the diffusion coefficient $D$ is obtained from the relation $\text{MSD} = 6D\Delta t$. 

We also calculate the end-to-end distance, $R_{ee} = |\vb{r}_{N} - \vb{r}_1|$, defined as the distance between the first and last beads on the polymer chain, and the persistence length $l_p$, obtained from the bond-bond correlation function $\left< \cos\theta(s) \right> = e^{-s / l_p}$, where $\theta(s)$ is the angle between tangent vectors of the chain separated by contour length $s$. The persistence length and end-to-end distance are not calculated for dendrimers, as these quantities are not well defined, and are evaluated only for the backbone in brush polymers and for a single arm in star polymers. The end-to-end distance is also not calculated for ring polymers. 

Additionally, we calculate the polymer form factor and the radial distribution function. The form factor is given by the isotropic intra-polymer structure factor~\cite{chen1986small,lindner2024neutrons}:
\begin{equation}
    P(q) = \left< \frac{\sin{(q |\vb{r}_i - \vb{r}_j|)}}{q |\vb{r}_i - \vb{r}_j|} \right>_{i,j}
\end{equation}
where $q$ is the magnitude of the scattering vector, and $\left< \dots \right>_{i,j}$ denotes the average over all bead pairs on the polymer. The radial distribution function is defined as
\begin{equation}
    g(r) = \left< \delta(|\vb{r}_i - \vb{r}_j| - r) \right>_{i,j}
\end{equation}
where $\delta$ is the Dirac delta function.

Finally, the human tool and report tool set are straightforward: the human tool allows the agent to receive natural-language feedback from the user, while the report tool reads and writes markdown files to the local directory.

%------------------------------
\begin{figure*}[!t]
    \centering
    \includegraphics[width=1\linewidth]{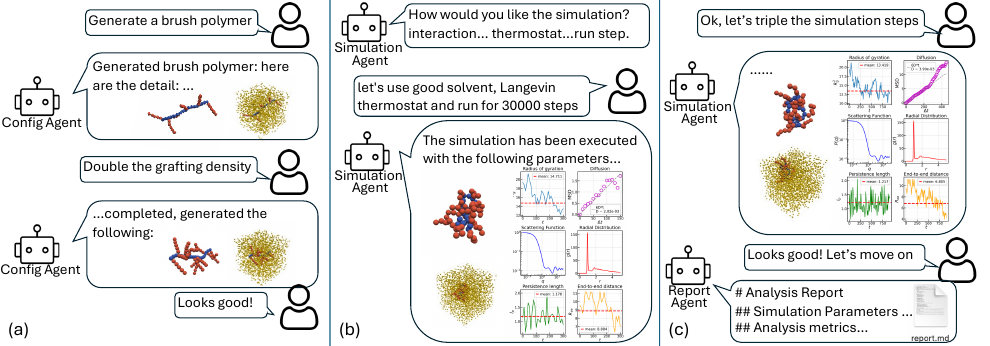}
    \caption{Example of ToPolyAgent assisting in a brush polymer simulation under interactive mode. Detailed text outputs from the agents are omitted for brevity. (a) The Config Agent and the user discuss the system configuration. (b) The Simulation Agent performs the MD simulation and presents the final system configuration and analysis results. (c) The user requests the Simulation Agent to extend the simulation, and the Report Agent compiles and presents a comprehensive report}
    \label{fig:interactive_brush}
\end{figure*}
%------------------------------

\subsection{Molecular Dynamics}
\label{ssec:Molecular Dynamics}

The main functionality of ToPolyAgent is to perform MD simulations of coarse-grained polymer models, in which the polymer is represented by a number of beads connected in different configurations. The simulations are carried out using LAMMPS. A truncated and shifted Lennard-Jones (LJ) potential is applied between every pair of beads~\cite{allen2017computer}:

\begin{equation}
\begin{aligned}
U^{ab}_{\mathrm{LJ}}(r) &= 
4\epsilon_{ab} \!\left[ \left( \frac{\sigma}{r} \right)^{12} - \left( \frac{\sigma}{r} \right)^{6} \right. \\[4pt]
&\quad \left. - \left( \frac{\sigma}{r_c} \right)^{12} + \left( \frac{\sigma}{r_c} \right)^{6} \right], \quad r \le r_c
\end{aligned}
\end{equation}

\noindent where $r$ is the distance between the two beads, $r_c=2.5\;\sigma$ is the cutoff distance, $\sigma$ is the Lennard-Jones characteristic length, and $\epsilon_{ab}$ is the interaction parameter that depends on the bead types. In our simulations of polymer–solvent mixture systems, $\epsilon_{pp}$, $\epsilon_{ss}$, and $\epsilon_{sp}$ correspond to polymer–polymer, solvent–solvent, and polymer–solvent interactions, respectively.

We use the finite extensible nonlinear elastic (FENE) bond potential for the connected beads on the polymer, given by~\cite{kremer1990dynamics}:

\begin{equation}
\begin{aligned}
    U_{\mathrm{FENE}}(r) &= -\frac{1}{2} K R_0^2 \ln{\left[ 1 - \left( \frac{r}{R_0} \right)^2 \right]} \\
    &\quad + 4\epsilon \left[ \left( \frac{\sigma}{r} \right)^{12} - \left( \frac{\sigma}{r} \right)^{6} \right] + \epsilon
\end{aligned}
\end{equation}

\noindent where $K$ is the spring constant, $R_0$ is the maximum bond extension, $\epsilon$ is the interaction strength, and $\sigma$ is the Lennard-Jones characteristic length. In the simulations, we fix $K = 30$, $R_0 = 1.5$, and $\epsilon = 1.0$ without loss of generality, thereby maintaining a manageable number of tuning parameters.

%In addition, to impart chain stiffness, a bending potential is applied to adjacent bonds:
%\begin{equation}
%U_{\mathrm{bend}}(\theta) = K_{\mathrm{bend}} \left( 1 - \cos (\theta) \right)
%\end{equation}
%\noindent where $K_{\mathrm{bend}}$ is the bending constant, and $\cos(\theta) = \hat{n}_i \cdot \hat{n}_{i+1}$, with $\theta$ being the angle between adjacent unit bond vectors.

We implement both the Langevin and Nosé–Hoover thermostats in the simulations. The equation of motion for the Langevin thermostat is given by~\cite{brunger1984stochastic,dunweg1991brownian}:
\begin{equation}
    m_i \frac{d^2 \vb{r}_i}{dt^2} = \vb{F}_i - \gamma m_i \frac{d \vb{r}_i}{dt} + \vb{R}_i(t)
\end{equation}
where $m_i$, $\vb{r}_i$, and $\vb{F}_i$ denote the mass, position, and interaction force of particle $i$, respectively. $\gamma$ is the friction coefficient, and $\vb{R}_i(t)$ is a Gaussian-distributed random force with zero mean and variance $\langle \vb{R}_i(t) \cdot \vb{R}_i(t') \rangle = 6 m_i \gamma k_B T \delta(t - t')$, where $k_B T$ is the target temperature and $\delta$ is the Dirac delta function. In our simulations, we set $m_i = 1$ for all particles, with a friction coefficient $\gamma = 1$ and target temperature $T = 1$ when using the Langevin thermostat.

The Nosé–Hoover thermostat maintains a constant temperature by introducing a fictitious degree of freedom that couples the system to a heat bath, thereby generating an NVT ensemble. The equations of motion are given by~\cite{evans1985nose,tuckerman2006liouville}:
\begin{equation}
\begin{aligned}
     m_i \frac{d^2 \vb{r}_i}{dt^2} &= \vb{F}_i - \xi m_i \frac{d \vb{r}_i}{dt}, \\
    \frac{d \xi}{dt} &= \frac{1}{Q} \left( \sum_i m_i \left( \frac{d \vb{r}_i}{dt} \right)^2 - g k_B T \right)
\end{aligned}
\end{equation}
where $\xi$ is the time-dependent friction coefficient, $Q$ is the fictitious mass of the thermostat, and $g$ is the number of degrees of freedom. In our simulations, the Nosé–Hoover thermostat is implemented using LAMMPS’ \texttt{fix nvt} command, with $Q = 0.1g$.

\section{Results}
\label{sec:result}

We begin by demonstrating examples of using ToPolyAgent to perform MD simulations of topological polymers with various architectures, including the interactive mode, which allows revisions during the process, and the autonomous mode, which executes the workflow to completion. We then showcase ToPolyAgent as a research assistant, capable of running multiple MD simulations through natural language interaction and compiling research findings for two representative cases: linear polymers and brush polymers.

\subsection{Interactive Mode}
In the interactive mode, ToPolyAgent engages with the user throughout the entire simulation process. We present two examples: one for a brush polymer, where multiple revisions are made based on user requests, and another for a star polymer, where the simulation proceeds in a streamlined manner.

Fig.~\ref{fig:interactive_brush} illustrates the workflow of the dialogue between the user and ToPolyAgent during the simulation of a brush polymer in solvent. Beginning with the generation of the brush polymer configuration and solvent packing shown in Fig.~\ref{fig:interactive_brush}(a), the Config Agent first produces a brush polymer using default settings. The user then provides feedback requesting an increase in grafting density. The Config Agent interprets this feedback, retrieves the previously used parameters for brush polymer generation from memory, and re-runs the configuration tool to adjust only the grafting density while keeping other parameters unchanged.

\begin{figure}[!t]
    \centering
    \includegraphics[width=1\linewidth]{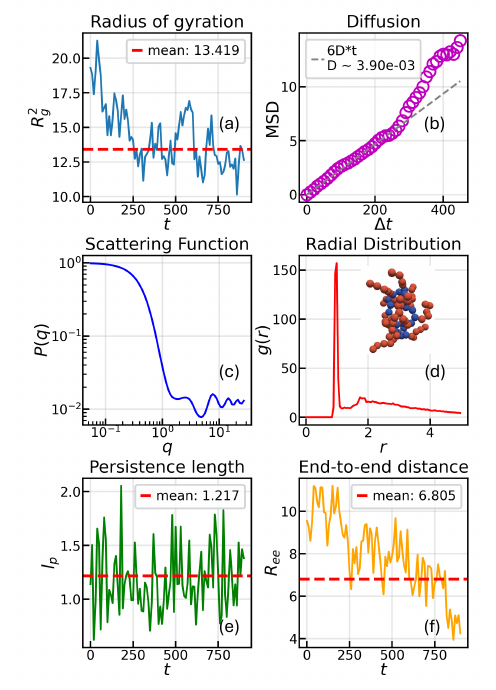}
    \caption{Conformation analysis of a brush polymer generated by ToPolyAgent. The system parameters are $N_b = 20$, $\sigma_g = 0.6$, $N_s = 5$, $n_s = 0.3$, $\epsilon_{pp} = \epsilon_{ss} = 0.3$, and $\epsilon_{sp} = 1.5$, with a simulation length of 100,000 steps under a Langevin thermostat. (a) Radius of gyration analysis. (b) Mean square displacement (MSD) analysis for diffusivity $D$ fitting. (c) Scattering function, or form factor $P(q)$. (d) Pair distribution function $g(r)$ with a snapshot of the polymer configuration inset. (e) Persistence length analysis. (f) End-to-end distance analysis. MSD, $P(q)$, and $g(r)$, as well as the mean values of $R_g^2$, $l_p$, and $R_{ee}$, are calculated using the second half of the simulation data.} 
    \label{fig:conformation_analysis_brush}
\end{figure}

After the user approves the generated system configuration, the Simulation Agent takes over, as shown in Fig.~\ref{fig:interactive_brush}(b). It begins by asking the user for additional specifications of the simulation parameters, then performs the MD simulation and corresponding polymer conformation analysis using the simulation tools, presenting the final system configuration and analysis results to the user. Subsequently, as shown in Fig.~\ref{fig:interactive_brush}(c), the user requests the Simulation Agent to extend the simulation to a longer run, which the agent executes accordingly. Upon final approval from the user, the Report Agent compiles a comprehensive report containing the simulation context, parameters, and conformation analysis results, with plots embedded in a markdown file.

The final simulation corresponds to a brush polymer with backbone length $N_b = 20$, grafting density $\sigma_g = 0.6$, and side chain length $N_s = 5$, within a simulation box of size $B = 20$ and solvent number density $n_s = 0.3$. The interaction parameters are $\epsilon_{pp} = \epsilon_{ss} = 0.3$ and $\epsilon_{sp} = 1.5$. The system is simulated using a Langevin thermostat for 90,000 steps with a time step of $dt = 0.01$. The complete conformation analysis results for the final run are shown in Fig.~\ref{fig:conformation_analysis_brush}.

\begin{figure}[!t]
    \centering
    \includegraphics[width=\linewidth]{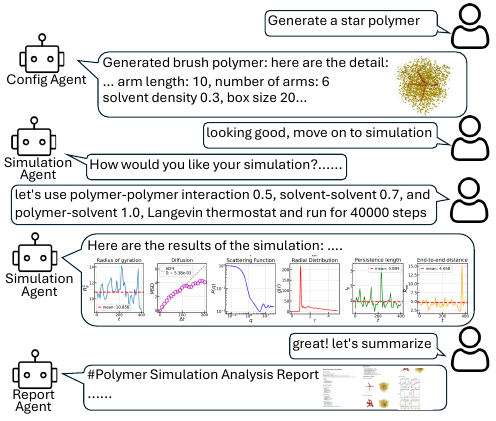}
    \caption{Example of ToPolyAgent assisting in a star polymer simulation under interactive mode.}
    \label{fig:interactive_star}
\end{figure}

Finally, in Fig.~\ref{fig:interactive_star}, we present an example of agent–user communication in which the user does not request any revisions during the process. The Config Agent selects its own parameters according to the system prompt defined in the tool’s docstring and generates a system consisting of a star polymer with arm length $N_a$ and number of arms $m = 6$ in a simulation box of size $B = 20$, filled with solvent at a number density of $n_s = 0.2$. The Simulation Agent then takes over and runs the simulation using user-specified parameters, including $\epsilon_{pp} = 0.5$, $\epsilon_{ss} = 0.7$, and $\epsilon_{sp} = 1.0$, for 40,000 steps under a Langevin thermostat. Finally, the Report Agent compiles the results and generates a markdown report.

\subsection{Autonomous Mode}
In the autonomous mode, ToPolyAgent automatically performs all stages of the topological polymer simulation and provides the user with the final reports. Compared with the interactive mode, which allows for detailed user tuning throughout the process, the autonomous mode streamlines the entire workflow and benefits from a more detailed initial prompt, enabling both the configuration generation and MD simulation steps to reflect user-specified parameters.

\begin{figure}[!h]
    \centering
    \includegraphics[width=\linewidth]{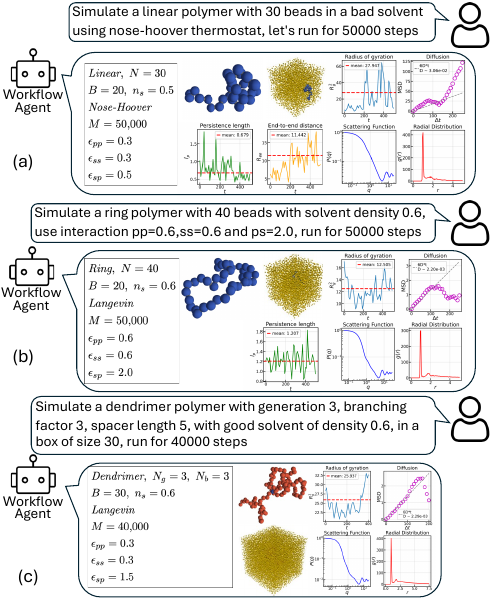}
    \caption{Examples of ToPolyAgent operating in autonomous mode. (a) Simulation of a linear polymer using the Nosé–Hoover thermostat. (b) Simulation of a ring polymer with specified interaction parameters. (c) Simulation of a dendrimer with a specified simulation box size.}
    \label{fig:autonomous_all}
\end{figure}

Fig.~\ref{fig:autonomous_all} shows three examples of ToPolyAgent operating in autonomous mode, demonstrating simulations of a linear polymer, a ring polymer, and a dendrimer. In Fig.~\ref{fig:autonomous_all}(a), the user instructs ToPolyAgent to simulate a linear polymer, specifying the polymer length, thermostat type (Nosé–Hoover), and a run length of 50,000 steps. ToPolyAgent correctly applies the user-specified parameters, performs the simulation, and provides the corresponding analysis results. In Fig.~\ref{fig:autonomous_all}(b), the user requests a ring polymer simulation, specifying the polymer size, solvent density, three interaction parameters for polymer–solvent interactions, and the total number of simulation steps. The agent accurately assigns these parameters and executes the simulation. The conformation analysis excludes the end-to-end distance, as it is not defined for ring polymers. Finally, in Fig.~\ref{fig:autonomous_all}(c), the agent receives detailed instructions for generating a dendrimer configuration, including solvent properties, simulation box size, and simulation length, and successfully performs the simulation using the correct parameter values. For dendrimers, since both the end-to-end distance $R_{ee}$ and persistence length $l_p$ are not well defined, only the other conformational characteristics are presented.

With the autonomous mode, ToPolyAgent can serve as a self-contained module within a larger system in which individual components communicate with one another through natural language, similar to human interaction. Despite relying on language-based inputs and outputs, ToPolyAgent employs rigorous scientific tools internally, producing more reliable and reproducible results than commercially available, general-purpose AI chatbots.

\subsection{ToPolyAgent as Research Assistant}

\begin{figure}[!t]
    \centering
    \includegraphics[width=1.0\linewidth]{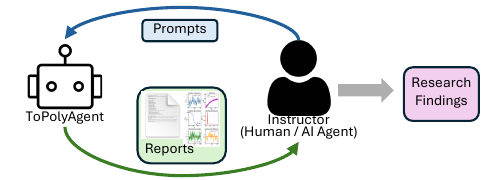}
    \caption{Illustration of ToPolyAgent acting as a research assistant specialized in coarse-grained simulations of topological polymers. The instructor can be either a human or another AI agent.}
    \label{fig:instructed_agent}
\end{figure}

Similar to a research assistant who conducts experiments and compiles reports based on instructions from an advisor, ToPolyAgent can play the same role --- translating natural-language instructions into simulation results and reporting them back to the instructor. These results can then contribute to new research findings. Fig.~\ref{fig:instructed_agent} illustrates this process. More importantly, because ToPolyAgent accepts natural-language input, the instructor does not necessarily have to be human --- it could also be another AI agent. To demonstrate the potential of ToPolyAgent as an AI research assistant, we present two example use cases: one involving a linear polymer and another involving a brush polymer.

For the linear polymer, we use ToPolyAgent to study the effect of solvent quality on conformational changes, particularly on the polymer’s radius of gyration. Although the simulation module supports explicit solvent particles, we simplify the study by using an implicit solvent model, where solvent effects are represented by the polymer–polymer interaction parameter $\epsilon_{pp}$, and the explicit solvent particle density is set to zero. We then provide ToPolyAgent with a series of prompt variations to perform simulations of linear polymers with different values of $\epsilon_{pp}$.

\smallskip
\fbox{%
\begin{minipage}{0.9\linewidth}
\textbf{Prompt:} Simulate a linear polymer consisting of 40 beads without solvent, using a polymer-polymer interaction parameter pp=\{epsilon\_pp\}. Set the box size to 40, apply a Langevin thermostat, and run the simulation for 100,000 steps
\end{minipage}%
}
\smallskip

\begin{figure}[!t]
    \centering
    \includegraphics[width=\linewidth]{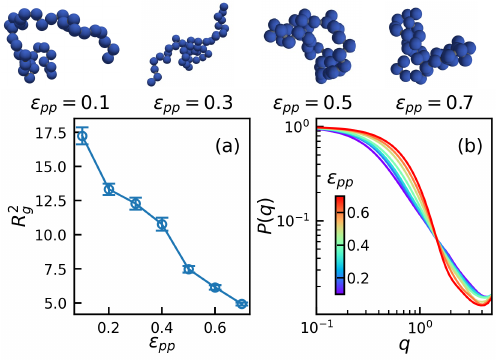}
    \caption{ToPolyAgent-assisted study (autonomous mode) of linear polymer conformation with implicit solvent. (a) Radius of gyration as a function of the polymer–polymer interaction parameter $\epsilon_{pp}$. (b) Polymer form factor $P(q)$ for different values of $\epsilon_{pp}$. Snapshots of polymer conformations at corresponding $\epsilon_{pp}$ values are shown at the top.}
    \label{fig:implicit_solvent}
\end{figure}

We then compile all simulation results to investigate the effect of solvent quality --- quantified by the polymer-polymer interaction parameter $\epsilon_{pp}$ --- on the size of the polymer chain. As shown in Fig.~\ref{fig:implicit_solvent}, the polymer chain collapses as the interaction strength increases. Fig.~\ref{fig:implicit_solvent}(a) shows the squared radius of gyration $R_g^2$ as a function of $\epsilon_{pp}$. As expected, increasing attractive interactions lead to a reduction in overall chain size, resulting in smaller $R_g^2$ values. Moreover, Fig.~\ref{fig:implicit_solvent}(b) characterizes the chain conformation through the slope of the form factor $P(q)$. The slope of $P(q)$ versus $q$ in the log–log scale increases with $\epsilon_{pp}$, indicating a transition from a self-avoiding walk at low $\epsilon_{pp}$, to an ideal random walk at intermediate $\epsilon_{pp}$, and finally to a collapsed globule at high $\epsilon_{pp}$, resembling a fuzzy sphere.

For the brush polymer, we use ToPolyAgent to study the effect of grafting density on the bending stiffness, characterized by the persistence length of the backbone. Similar to the linear polymer study, we provide ToPolyAgent with a series of variations of the following prompt to simulate brush polymers in a solvent bath with different grafting densities:

\smallskip
\fbox{%
\begin{minipage}{0.9\linewidth}
\textbf{Prompt:} Simulate a brush polymer with 20 beads on the backbone, grafting density \{grafting\_density\}, and side chain length 5, in a good solvent with number density 0.2. Use a box size of 30, apply a Langevin thermostat, and run the simulation for 100000 steps
\end{minipage}%
}
\smallskip

\begin{figure}[!t]
    \centering
    \includegraphics[width=\linewidth]{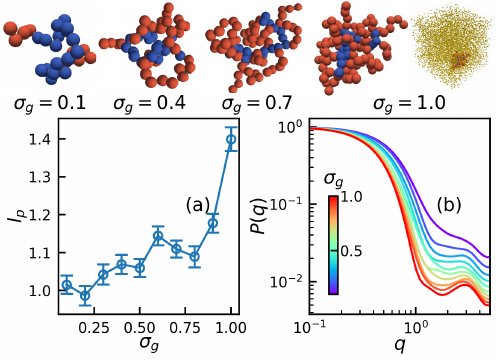}
    \caption{ToPolyAgent-assisted study of the effect of grafting density $\sigma_g$ on brush polymer conformation. (a) Persistence length $l_p$ of the polymer backbone as a function of grafting density. (b) Polymer form factor $P(q)$ for different values of $\sigma_g$. Snapshots of polymers with varying grafting densities are shown at the top, with the full system including solvent displayed for $\sigma_g = 1$.}
    \label{fig:brush_persistence}
\end{figure}

Fig.~\ref{fig:brush_persistence} shows the effect of grafting density $\sigma_g$ on the conformation of the brush polymer. As the grafting density increases, the region near the backbone becomes increasingly crowded, resulting in reduced backbone flexibility. As shown in Fig.~\ref{fig:brush_persistence}(a), the persistence length $l_p$ of the backbone chain increases with $\sigma_g$. Meanwhile, the form factor of the entire polymer, shown in Fig.~\ref{fig:brush_persistence}(b), decreases more rapidly at higher grafting densities, indicating a transition from an extended polymer-like structure to a more compact, sphere-like morphology due to the dense side chains. This conformational transformation is illustrated by the snapshots of polymers with different grafting densities shown at the top of Fig.~\ref{fig:brush_persistence}.

\section{Summary}
\label{sec:summary}
% what we've done
In this work, we introduce ToPolyAgent, a multi-agent AI system specialized in MD simulations of coarse-grained topological polymers. The ToPolyAgent system consists of four LLM-powered agents, each assigned specific tasks and equipped with dedicated tools to execute different stages of the simulation workflow. These agents are: the Config Agent, which generates the initial system configuration including the polymer and solvent; the Simulation Agent, which performs MD simulations and conducts conformational analyses of the results; the Report Agent, which collects logs and generated files from the simulation workflow and compiles a report in markdown format; and the Workflow Agent, which integrates the functions of the Config Agent and Simulation Agent.

ToPolyAgent operates in two modes: interactive and autonomous. In the interactive mode, the Config Agent, Simulation Agent, and Report Agent are activated, with the Config and Simulation Agents equipped with a human-interaction tool that enables real-time communication with the user for revision and clarification. In the autonomous mode, the Workflow Agent and Report Agent automatically produce the final report directly from the user’s initial prompt, providing a more streamlined simulation process.

We present examples of both interactive and autonomous modes, covering five common types of topological polymers: linear polymers, ring polymers, brush polymers, star polymers, and dendrimers. Finally, we demonstrate the potential of ToPolyAgent as a research assistant capable of handling more complex research tasks and automating parts of the scientific discovery process.

% what is the significance 
Building on the results demonstrated in this work, ToPolyAgent represents a new direction in computational materials science by seamlessly integrating LLM-enabled agentic AI systems with rigorous simulation tools. This integration not only extends the capabilities of existing AI frameworks but also enhances the practical utility of the simulation tools developed. By enabling researchers to interact with the system through natural language, ToPolyAgent significantly lowers the technical barriers to configuring and executing complex polymer simulations, thereby making advanced computational tools accessible to a broader range of users.

More importantly, ToPolyAgent’s potential as a research assistant paves the way for autonomous scientific research in computational materials science. By combining ToPolyAgent with other specialized research agents, it becomes possible to construct a virtual research group composed entirely of AI agents with complementary expertise. For example, a discovery agent could identify research gaps addressable through topological polymer simulations by reviewing relevant literature; a planning agent could design a research plan involving specific sets of simulations, which are then executed by ToPolyAgent; and a writing agent could compile a well-structured research paper based on the simulation results. Human researchers would oversee and refine this process---analogous to ToPolyAgent’s interactive mode---ensuring alignment with scientific objectives.

By enabling scalable, collaborative, and automated research workflows, ToPolyAgent has the potential to accelerate discovery in polymer science and beyond, constrained only by available computational resources.

% what are the futhre directions
Looking forward, the development of ToPolyAgent opens several promising avenues for advancing computational materials science and AI-driven research. First, extending the system’s capabilities to handle more complex polymer systems --- such as copolymers --- and integrating additional molecular simulation tools such as CHARMM~\cite{brooks2009charmm} and CHARMM-GUI~\cite{jo2017charmm}, which also support coarse-grained modeling, could broaden its applicability to real-world materials design challenges. Second, incorporating advanced analysis methods, such as machine learning models for interpreting polymer properties from simulation data via simulation-based inference~\cite{ding2025sphere,ding2025colloids,ding2024mechanical,ding2025charge,ding2025ladder,ding2025deciphering,tung2025scattering,tung2025insights}, could further streamline the research process by providing predictive insights alongside simulation results. Third, integrating LLM-based AI agents with simulation frameworks supporting Monte Carlo methods --- such as GOMC~\cite{nejahi2019gomc}, SPARKS~\cite{mitchell2023parallel} and HOOMD-blue~\cite{anderson2020hoomd}, would further enhance the system’s versatility. Finally, expanding ToPolyAgent’s interoperability within a modular AI research ecosystem represents an important future direction. For example, coupling ToPolyAgent with agents specialized in experimental data analysis could enable direct comparison between simulated and experimental results, fostering deeper integration between computation and experiment.

\section*{Data Availability}
The code for this work is available at the GitHub repository \url{https://github.com/ljding94/ToPolyAgent}

\section*{Author Contributions}
LD, JMC and CD conceived this work; LD designed, implemented, and tested the agentic AI system; LD, JMC and CD wrote and edited the manuscript.

\section*{Acknowledgment}
This research was performed at the Spallation Neutron Source, which is a DOE Office of Science User Facilities operated by Oak Ridge National Laboratory. This research was sponsored by the Laboratory Directed Research and Development Program of Oak Ridge National Laboratory, managed by UT-Battelle, LLC, for the US DOE. Portions of the computational aspect of this research were supported by the Center for Nanophase Materials Sciences (CNMS), which is a U.S. Department of Energy Office of Science User Facility at Oak Ridge National Laboratory. 

% The \nocite command causes all entries in a bibliography to be printed out
% whether or not they are actually referenced in the text. This is appropriate
% for the sample file to show the different styles of references, but authors
% most likely will not want to use it.
%\nocite{*}

%\clearpage
%\section*{References}
\bibliography{reference}% Produces the bibliography via BibTeX.

%\clearpage
\onecolumngrid
\clearpage
\appendix

\end{document}